\documentclass{article}
\usepackage{ijcai25} 
\usepackage{times}
\usepackage{soul}
\usepackage{subfigure}
\usepackage{multirow}

\usepackage{url}
\usepackage[hidelinks]{hyperref}
\usepackage[utf8]{inputenc}
\usepackage[small]{caption}
\usepackage{graphicx}
\usepackage{amsmath}
\usepackage{amsthm}
\usepackage{booktabs}
\usepackage{algorithm}
\usepackage{algorithmic}
\usepackage[switch]{lineno} 
\usepackage{amssymb}
\usepackage{enumitem}
\usepackage{bm}   

\title{An Active Diffusion Neural Network for Graphs}

\author{
    Mengying Jiang
    \affiliations
    \emails
    mengyingjianggdut@foxmail.com
}
 
\begin{document}

\maketitle

\begin{abstract}
The analogy to heat diffusion has enhanced our understanding of information flow in graphs and inspired the development of Graph Neural Networks (GNNs). However, most diffusion-based GNNs emulate passive heat diffusion, which still suffers from over-smoothing and limits their ability to capture global graph information. Inspired by the heat death of the universe—which posits that energy distribution becomes uniform over time in a closed system—we recognize that, without external input, node representations in a graph converge to identical feature vectors as diffusion progresses.
To address this issue, we propose the Active Diffusion-based Graph Neural Network (ADGNN). ADGNN achieves active diffusion by integrating multiple external information sources that dynamically influence the diffusion process, effectively overcoming the over-smoothing problem. Furthermore, our approach realizes true infinite diffusion by directly calculating the closed-form solution of the active diffusion iterative formula. This allows nodes to preserve their unique characteristics while efficiently gaining comprehensive insights into the graph's global structure.
We evaluate ADGNN against several state-of-the-art GNN models across various graph tasks. The results demonstrate that ADGNN significantly improves both accuracy and efficiency, highlighting its effectiveness in capturing global graph information and maintaining node distinctiveness.

\end{abstract}

\section{Introduction}

With the increasing prevalence of graph-structured data, GNNs have gained growing attention for their ability to capture the complex interactions between nodes \cite{PAMI3}. 
In recent years, numerous powerful GNNs have emerged \cite{{GCN+chapter2017},{GraphSage+chapter2017},{GAT+chapter2018},{GIN+chapter2019},{MixHop}}, showcasing exceptional performance in handling graph-structured data \cite{nips2}. 
However, current GNNs still face significant hurdles in real-world applications. For instance, common noise and outliers in graph data can lead to incorrect connections, disrupting model training and inference. In specific domains, such as the Internet of Things and social platforms, the constructed graph-structured data may exhibit heterophily \cite{nlgnn}, resulting in more complex relationships between nodes. In these cases, traditional GNN models often fail to achieve the desired performance \cite{PAMI5}. Additionally, as large-scale graph datasets continue to grow, conventional GNN architectures may struggle to process information efficiently, imposing higher demands on model scalability and real-time responsiveness \cite{DIFFORMER}.

Recently, modeling information propagation between nodes using heat diffusion has emerged as a promising approach in graph representation learning for complex scenarios. For example, GDC \cite{GDC} transforms the learning process on graphs into a continuous diffusion process, treating GNNs as discrete approximations of partial differential equations (PDE). GRAND++ \cite{GRAND++} enhances the diffusion process by introducing a source term, successfully addressing the over-smoothing problem, and building a very deep model. GREAD \cite{GREAD} introduces a reaction-diffusion-based method, providing a more flexible framework for modeling complex interactions in graph data. DIFFormer \cite{DIFFORMER} designs an energy-driven diffusion model based on transformers, ensuring each diffusion step moves towards a global optimum. HiD-Net \cite{HiD-Net} proposes a general diffusion framework with fidelity terms, formally defining the intrinsic connections between the diffusion process and various GNN architectures.

Modeling information propagation on graphs as heat diffusion offers several advantages for GNNs \cite{DIFFUSION0}. For instance, by adjusting the diffusion step size, models can effectively control the speed of information propagation, thereby alleviating the over-smoothing problem \cite{DIFFORMER}. This helps GNNs achieve a more comprehensive representation and understanding of graph data \cite{DIFFUSION0}.
However, existing diffusion-based GNNs still face significant limitations. Primarily, they typically simulate passive heat diffusion, which causes node features to eventually converge to the same vector as diffusion progresses \cite{GRAND++}. This inherent over-smoothing makes it difficult for these models to capture and integrate global graph information, thereby limiting their generalization capabilities. Moreover, the reliance on iterative diffusion steps to capture long-range dependencies introduces substantial computational overhead, making it challenging for traditional GNN architectures to efficiently handle large-scale graph data.

In this work, we present the ADGNN to address the limitations of existing diffusion-based GNNs. ADGNN constructs a novel active diffusion mechanism by integrating three significant source terms: ego embeddings, boundary detection, and anomaly detection.
These source terms continuously inject unique and valuable information into nodes during the diffusion process, preventing node features from converging to identical feature vectors.
This approach not only effectively addresses the over-smoothing issue, but also enables the model to capture richer and more comprehensive graph information.
Additionally, we calculate a closed-form solution for the active diffusion formula, enabling infinite diffusion iterations to be performed in a single computational step. 
This strategy allows the model to integrate global structural information efficiently.
We summarize the major practical advantages of ADGNN below:

\begin{itemize}[left=0pt]
  \item \textbf{Active Diffusion Mechanism}: First GNN to incorporate active heat diffusion with multiple source terms, preventing over-smoothing and ensuring distinct node features.
  
  \item \textbf{Three Source Terms Integration}: Combines three meaningful source terms (ego embeddings, boundary detection, and anomaly detection) to enrich the node representations.

  \item \textbf{Closed-Form Solution}: Enables infinite diffusion iterations in one step, enhancing efficiency and scalability.

  \item \textbf{Superior Performance}: Extensive experiments show that ADGNN outperforms most existing methods in accuracy and efficiency on node classification tasks across diverse graph scales and homophily levels.

  
\end{itemize}

\section{Related Work}
\label{Preliminaries}

\subsection{Node Embedding Update Method}
\label{GNNs}
 
In GNNs, a generalized update formula for node embeddings can be expressed as:
\begin{equation}
\begin{aligned}
{\bf h}^{(k)}_u = f \left( {\bf h}^{(k-1)}_u, \{{\bf h}^{(k-1)}_v : v \in \mathcal N_{u}\} \right),\\
\quad \text{s.t.}\quad {\bf h}^{(0)}_u = {\bf x}_u,\quad k\geq 0,
\end{aligned}
\label{a1}
\end{equation}
where the function \(f\) is applied repeatedly for \(K\) total rounds to integrate features over longer distances.
${\bf h}^{(k)}_u$ denotes the embedding of the node \(u\) at round \(k\), derived from both its own and its neighbors' embeddings from the previous round \cite{hhgcn}.
$\mathcal N_{u}$ represents the set of neighboring nodes for node $u$.
By adjusting the value of \(K\), GNNs are able to capture neighborhood information at various distances, thereby adapting to different task requirements \cite{GraphSage+chapter2017}.
However, this widely-used embedding update method is typically effective for homophilic graphs but struggles with heterophilic graphs, which are common in real-world scenarios \cite{GREAD}. Additionally, repeated rounds of feature propagation and aggregation can lead to the loss of node-specific information, causing the model to suffer from over-smoothing \cite{GRAND++}.
To address these challenges, numerous GNNs have been proposed, with diffusion-based GNNs performing particularly well \cite{GREAD}.

\subsection{Passive Diffusion on Graphs}

The information diffusion process on graphs is inspired by the heat diffusion principle using the graph laplacian, generating instance representations through state evolution \cite{DIFFORMER}. A key feature is the anisotropic diffusion mechanism, which dynamically adjusts diffusion weights to enhance effective information propagation among instances \cite{DIFFORMER}.
We represent the state of instance \( u \) at time \( t \) by the vector \( {\bf z}_u^{(t)} \). The state evolves according to the PDE:
\begin{equation}
\begin{aligned}
    \frac{\partial {\bm{Z}}^{(t)}}{\partial t}\hspace{-1mm} = \hspace{-1mm}\nabla^* \big( \bm{S}(\bm{Z}^{(t)}, t) \odot \nabla \bm{Z}^{(t)} \big),
    ~\text{s.t.} ~\bm{Z}^{(0)} \hspace{-1mm}= \hspace{-1mm}[{\bf x}_u]_{u=1}^N,
    \label{eq:pde_compressed}
    \end{aligned}
\end{equation}
where \( \odot \) denotes the Hadamard product, The matrix \(\bm{Z}^{(t)} = [{\bf z}_u^{(t)}]_{u=1}^N \in \mathbb{R}^{N \times d} \) represents the state matrix of \( N \) instances at time \( t \), and \( \bm{S}(\bm{Z}^{(t)}, t) \) is the anisotropic diffusion coefficient matrix, adjusting diffusion rates based on instance states and adjacency \cite{DIFFORMER}. The operators \( \nabla \) and \( \nabla^* \) represent the gradient and divergence, respectively, capturing state differences and aggregating weighted information flow.

To discretize the PDE, the explicit Euler method with step size \( \tau \) is applied, yielding the state update:
\begin{equation}
\begin{aligned}
    {\bf z}_u^{(k+1)} &= {\bf z}_u^{(k)} - \tau \sum_{v=1}^N S_{uv}^{(k)} \big({\bf z}_u^{(k)} - {\bf z}_v^{(k)}\big) \\
    &= \underbrace{\left(1 - \tau \sum_{v=1}^N S_{uv}^{(k)}\right) {\bf z}_u^{(k)}}_{\text{\normalsize State Conservation}}
    + \underbrace{\tau \sum_{v=1}^N S_{uv}^{(k)} {\bf z}_v^{(k)}}_{\text{\normalsize State Propagation}}.
    \label{eq:discrete_update_compressed}
\end{aligned}
\end{equation}
In matrix form, the update rule is:
\begin{equation}
    \bm{Z}^{(k+1)} = \big({\bf I} - \tau \bm{D}^{(k)} \big) \bm{Z}^{(k)} + \tau \bm{S}^{(k)} \bm{Z}^{(k)},
    \label{eq:matrix_update_compressed}
\end{equation}
where \( \bm{D}^{(k)} = \operatorname{diag}(\bm{S}^{(k)} {\bf 1}_N) \) is the degree matrix. Often, \( \bm{S}^{(k)} \) is row-normalized, simplifying the update to:
\begin{equation}
    \bm{Z}^{(k+1)} = (1 - \tau) \bm{Z}^{(k)} + \tau \bm{S}^{(k)} \bm{Z}^{(k)},
    \label{eq:normalized_update_compressed}
\end{equation}
with \( \tau \in (0,1) \). Adjusting \( \tau \) controls the diffusion speed, mitigating over-smoothing and enabling information to propagate over longer distances.
We refer to graph models where node representations rely solely on their own and their neighbors' previous states, without external information sources, as “passive diffusion-based GNNs.”
According to the heat death of the universe \cite{miralda2003dark}, in a closed system without external heat sources, energy distribution becomes uniform over time. As a result, these “passive diffusion-based GNNs” remain at risk of over-smoothing and struggle to capture the global structural information of the graph.

\section{The Proposed Active Diffusion-Based Graph Neural Networks}
\label{proposed method}

\begin{figure*}[t]
\centering
\includegraphics[width=6.5in,height=2.8in]{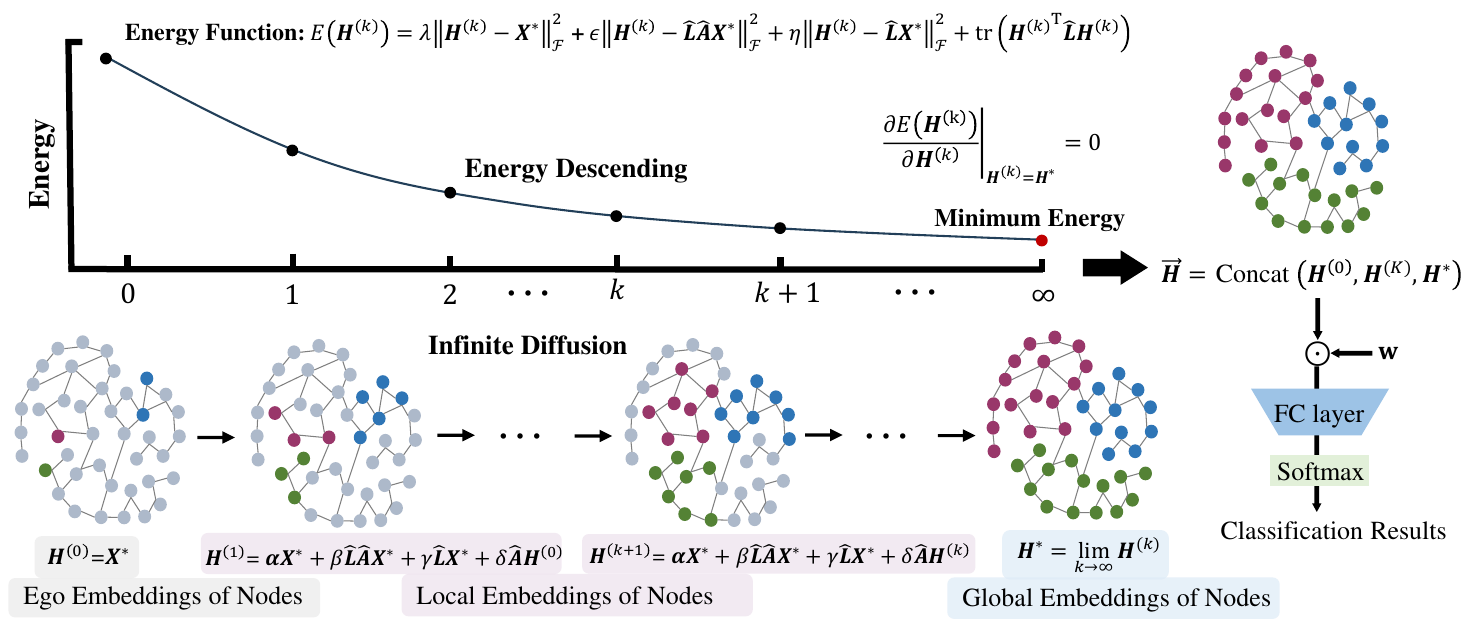}
\vspace{-2mm}
\caption{Flowchart of the proposed ADGNN. 
Initially, ADGNN computes the ego embeddings for nodes, ${\bm X}^{*}$, utilizing their initial features.
Through $K$ diffusion iterations, ADGNN achieves the local embeddings of nodes, ${\bm H}^{(K)}$, reflecting the local structural information of the graph. 
As the diffusion progresses, the energy of the graph decreases.
Upon completion of infinite diffusion, the diffusion output, ${\bm H}^{*}$, is obtained, which encapsulates the global information of the graph and serves as the global embeddings for the nodes. 
At this stage, the energy of the graph also reaches its minimum.
Ultimately, by integrating three diffusion scales of node embeddings (i.e. ${\bm H}^{0}$, ${\bm H}^{(K)}$, and ${\bm H}^{*}$), We derive a high-level node embedding for node classification.}
\label{flow}
\end{figure*}

Let \( \mathcal{G} = (\mathcal{V}, \mathcal{E}) \) be an undirected graph, where \( \mathcal{V} \) represents the set of nodes and \( \mathcal{E} \) denotes the set of edges. The node features are represented by \( \bm X = \left\{ \mathbf{x}_v \right\}_{v=1}^{N} \in \mathbb{R}^{N \times d} \), where \( N = |\mathcal{V}| \) is the number of nodes, and \( d \) indicates the dimensionality of the feature vector for each node. The adjacency matrix is denoted by \( \bm A \in \mathbb{R}^{N \times N} \). The symmetric normalized adjacency matrix is defined as \({\hat{\bm A}} = {\widetilde{\bm D}}^{-1/2}{\widetilde{\bm A}} {\widetilde{\bm D}}^{-1/2},\)
where \( {\widetilde{\bm A}} = \bm A + \bm I \), and \( {\widetilde{\bm A}} = \operatorname{diag}({\widetilde{\bm A}} \mathbf{1}_N) \) is the degree matrix corresponding to the adjacency matrix with self-loops.

In this work, we focus on the node classification task, which aims to learn a mapping \( f: \mathcal{V} \to \mathcal{Y} \), where \( \mathcal{Y} \) is the set of labels. Given a set of labeled nodes \( \mathcal{V}_{\text{labeled}} \) as the training data. Once trained, the learned mapping \( f \) is used to predict the labels of unlabeled nodes \( v \in \mathcal{V} \) and \( v \notin \mathcal{V}_{\text{labeled}} \), based on their feature representations.
The workflow of our ADGNN model is depicted in Fig. \ref{flow}.

\subsection{Active Diffusion on Graphs}
\label{Active Diffusion on Graphs}

To completely overcome the over-smoothing problem and understand the overall structure of the graph, we introduce three external information sources into the diffusion process, forming an active diffusion mechanism. First, we calculate the ego embeddings of the nodes as follows:
\begin{equation}
\bm X^{*} = \operatorname{ReLU}\left( \bm X \bm W' + \mathbf{b}' \right),
\label{ActiveD1}
\end{equation}
where \( \bm X \in \mathbb{R}^{N \times d} \) represents the initial features of nodes, \( \bm W' \in \mathbb{R}^{d \times d'} \) is a learnable weight matrix, and \( \mathbf{b}' \in \mathbb{R}^{1 \times d'} \) is the bias. The ego embeddings \( \bm X^{*} \in \mathbb{R}^{N \times d'} \) reflect the intrinsic attribute characteristics of each node.

Next, an active diffusion equation is proposed to update the node embeddings:
\begin{equation}\renewcommand{\arraystretch}{2.5}
\begin{aligned}
\bm H^{(k+1)} = &\underbrace{\alpha \bm X^{*}}_{\text{\normalsize  Term 1}} + \underbrace{\beta \hat{\bm L} {\hat{\bm A}} \bm X^{*}}_{\text{\normalsize  Term 2}} + \underbrace{\gamma {\hat{\bm L}} \bm X^{*}}_{\text{\normalsize Term 3}} + \delta {\hat{\bm A}} \bm H^{(k)},\\
&\quad \text{s.t.}\quad \bm H^{(0)} = \bm X^{*}, \, k \geq 0,
\end{aligned}
\label{ActiveD2}
\end{equation}
where \( \bm H^{(k)} \in \mathbb{R}^{N \times d'} \) represents the node embeddings after \( k \) diffusion iterations.
Terms 1, 2, and 3 are the source terms that form the active diffusion mechanism.
Specifically, term 1 is ego embeddings for nodes, ensuring that the current embeddings remain consistent with the intrinsic attributes of the nodes. 
Terms 2 and 3 are inspired by the Laplacian of Gaussian (LoG) and Laplacian operators \cite{LOG} used in image processing, which can be used for boundary detection and anomaly detection of graphs.

In images, the LoG operator first applies a Gaussian filter to smooth an image and then uses the Laplacian operator to detect edges \cite{LOG}. In graphs, \( {\hat{\bm A}} \) serves as the symmetric normalized adjacency matrix, and \( {\hat{\bm A}} {\bm X}^{*} \) acts as a low-pass filter that reduces noise in graph signal. The matrix \( {\hat{\bm L}} = {\hat{\bm D}} - {\hat{\bm A}} \) is the normalized Laplacian matrix, where \( {\hat{\bm D}} \) is the degree matrix of \( {\hat{\bm A}} \). Hence, \( {\hat{\bm L}} {\hat{\bm A}} \bm X^{*} \) performs a LoG operation on the graph, detecting boundaries and local variations. Term 3 applies the Laplacian operator directly on \( \bm X^{*} \) to preserve anomaly information.
These three source terms provide valuable information for each node throughout the diffusion iterations, ensuring that each node has a stable and distinct representation, even after infinite iterations. 
The parameters \( \alpha \), \( \beta \), \( \gamma \), and \( \delta \) are nonnegative tuning parameters representing the weights of the components. To ensure convergence and numerical stability, these parameters are constrained as \( 0 \leq \alpha, \beta, \gamma < 1 \), \( 0 < \delta < 1 \), and \( \alpha + \beta + \gamma + \delta = 1 \).

To enable ADGNN to capture the true global information, we aim to extend $k$ to infinity. However, infinitely stacking the node embedding update layer Eq.~\eqref{ActiveD2} is impractical. 
Instead, we directly compute the closed-form solution of this active diffusion equation.
The process is as follows: First, we analyze Eq.~\eqref{ActiveD2} inductively and derive the following iterative formula:
\begin{equation}
\begin{aligned}
\bm H^{(k+1)} =& \sum_{t=0}^{k} (\delta {\hat{\bm A}})^t \left( \alpha \bm X^{*} + \beta {\hat{\bm L}} {\hat{\bm A}} \bm X^{*} + \gamma {\hat{\bm L}} \bm X^{*} \right)\\
&+ (\delta{\hat{\bm A}})^{k+1} \bm X^{*}.
\label{ActiveD3}
\end{aligned}
\end{equation}
Next, we extend $k$ to infinity. Since the eigenvalues of \( {\hat{\bm A}} \) lie within \( \mathbb{R}[-1, 1] \), and \( 0 < \delta < 1 \), we obtain:
\begin{equation}
\lim_{k \to \infty} (\delta {\hat{\bm A}})^{k+1} = 0.
\label{ActiveD4}
\end{equation}
As \( k \) approaches infinity, the non-zero term in Eq. \eqref{ActiveD3} is calculated as follows: 
\vspace{-1mm}
\begin{equation}
\begin{aligned}
\bm H^{*} &=\lim_{k \to \infty} \sum_{t=0}^{k} (\delta {\hat{\bm A}})^t \left( \alpha \bm X^{*} + \beta {\hat{\bm L}} {\hat{\bm A}} \bm X^{*} + \gamma {\hat{\bm L}} \bm X^{*} \right) \\
&= \left( \bm I - \delta {\hat{\bm A}} \right)^{-1} \left( \alpha \bm X^{*} + \beta {\hat{\bm L}} {\hat{\bm A}} \bm X^{*} + \gamma {\hat{\bm L}} \bm X^{*} \right).
\label{ActiveD8}
\end{aligned}
\end{equation}
where the non-zero term $\bm H^{*}$ can be regarded as the node embeddings after infinite diffusion iterations, and also as the closed-form solution of Eq. \eqref{ActiveD2}.

The computation of the closed-form solution \( \bm H^{*} \) reveals several key advantages of our ADGNN over other passive diffusion-based GNNs:
1) Even after an infinite number of diffusion iterations, the nodes do not converge to the same feature vector, which demonstrates that our ADGNN effectively addresses the over-smoothing problem;
2) The number of iterations $k$ can be extended to infinity, confirming that our ADGNN is capable of capturing global structural information;
3) Global structural information is obtained in a single computation, highlighting the high efficiency of our ADGNN.
Herein, we call \(\bm H^{*} \) the global embeddings of nodes.

\subsection{Energy Function}
\label{Energy Function}
This subsection aims to validate the effectiveness of node global embeddings \( \bm H^{*} \).
We develop an energy function to measure the quality of the node representations as follows:
\vspace{-1mm}
\begin{align}
&\displaystyle{E}({\bm H}^{(k)})= \lambda \left\|{\bm H}^{(k)}\hspace{-0.7mm}-\hspace{-0.7mm}{\bm X}^{*}\right\|_{\mathcal{F}}^2+
\epsilon \left\|{\bm H}^{(k)}\hspace{-0.7mm}-\hspace{-0.7mm}{\hat {\bm L}}{\hat {\bm A}}{\bm X}^{*}\right\|_{\mathcal{F}}^2 \nonumber\\
&+ \eta\left\|{\bm H}^{(k)}\hspace{-0.7mm}-\hspace{-0.7mm}{\hat {\bm L}}{\bm X}^{*}\right\|_{\mathcal{F}}^2+ \frac{1}{2}\sum_{i, j=1}^{N} {\hat A}_{i,j} \left\|\mathbf{h}_i^{(k)}\hspace{-0.7mm}-\hspace{-0.7mm}\mathbf{h}_j^{(k)}\right\|_{\mathcal{F}}^2, \label{energy1}
\end{align} 
\vspace{-1mm}
where ${\bm H}^{(k)}$ represents the current representations of nodes.
$\lambda$, $\epsilon$, $\eta \geq 0$ serve as tuning parameters.
The first term on the right-hand side of the energy function represents the fitting constraint, indicating that the current representations of nodes should not deviate significantly from the ego embeddings.
The second and third terms represent other fitting constraints, suggesting ideal node representations should be sensitive to boundary and anomaly information within graphs.
The fourth term is the smoothing term, which encourages spatial continuity of the segmentation results and avoids overly fragmented segmentation.

By differentiating $\displaystyle{E}({\bm H}^{(k)})$ with respect to ${\bm H}^{(k)}$, we can obtain the optimal solution $\hat{\bm H}$, the configuration that minimizes the energy function for a given value of $\lambda$, $\epsilon$ and $\eta$.
To facilitate the calculation of $\hat{\bm H}$, the energy function needs to be transformed as follows:
\begin{align}
&\displaystyle{E}({\bm H}^{(k)})=\lambda \left\|{\bm H}^{(k)}\hspace{-0.7mm}-\hspace{-0.7mm}{\bm X}^{*}\right\|_{\mathcal{F}}^2+
\epsilon \left\|{\bm H}^{(k)}\hspace{-0.7mm}-\hspace{-0.7mm}{\hat {\bm L}}{\hat {\bm A}}{\bm X}^{*}\right\|_{\mathcal{F}}^2 \nonumber\\
&+ \eta \left\|{\bm H}^{(k)}\hspace{-0.7mm}-\hspace{-0.7mm}{\hat {\bm L}}{\bm X}^{*}\right\|_{\mathcal{F}}^2+\text{tr}\left({{\bm H}^{(k)}}^{\top}{\hat {\bm L}} {{\bm H}^{(k)}}\right). \label{energy3}
\end{align} 

Differentiating $\displaystyle{E}({\bm H}^{(k)})$ with respect to ${\bm H}^{(k)}$, we have: 
\begin{equation} 
\begin{aligned}
&\left.\frac{\partial\displaystyle{E}({\bm H}^{(k)})}{\partial {\bm H}^{(k)}}\right|_{{\bm H}^{(k)}=\hat{\bm H}} 
= \lambda\left( \hat{\bm H}-{\bm X}^*\right)\\&+\epsilon\left( \hat{\bm H}-{\hat {\bm L}}{\hat {\bm A}}{\bm X}^* \right)
+\eta\left( \hat{\bm H}-{\hat {\bm L}}{\bm X}^* \right)+{\hat {\bm L}}\hat{\bm H}. 
\label{energy44}
\end{aligned} 
\end{equation}

Consequently, the optimal solution $\hat{\bm H}$ that minimizes the value of the $\displaystyle{E}({\bf H}^{(k)})$ is calculated as:
\begin{equation}  
{\hat {\bm L}} \hspace{-0.8mm}=\hspace{-0.8mm}\left(\hspace{-0.6mm} {\bm I}\mbox{ -- }\frac{{\hat {\bm A}}}{\lambda\hspace{-0.5mm}+\hspace{-0.5mm}\epsilon\hspace{-0.5mm}+\hspace{-0.5mm}\eta\hspace{-0.5mm}+\hspace{-0.5mm}1}\hspace{-0.6mm}\right)^{\mbox{-}1}\hspace{-2.5mm} \left(\lambda{\bm X}^*\hspace{-0.4mm}+\hspace{-0.4mm}\epsilon{\hat {\bm L}}{\hat {\bm A}}{\bm X}^*\hspace{-0.4mm}+\hspace{-0.4mm}\eta{\hat {\bm L}} {\bm X}^*\hspace{-0.6mm}\right). 
\label{energy6}
\end{equation} 

Herein, the optimal solution $\hat{\bm H}$ presents an optimal state for node representations.
Obviously, the optimal solution $\hat{\bm H}$ of the energy function and the closed-form solution ${\bm H}^*$ in
Eq.~\eqref{ActiveD8} have similar expressions.
If we assume that:
\begin{equation}
\lambda=\frac{\alpha}{\delta}, \epsilon=\frac{\beta}{\delta} ~\text {and}~\eta = \frac{\gamma}{\delta},
\label{energy7}
\end{equation} 
the optimal solution $\hat{\bm H}$ is transformed as:
\begin{equation}
\begin{aligned}
\hat{\bm H} &=\left( {\bm I}\hspace{-0.2mm}-\hspace{-0.2mm} \delta \hat{\bm A} \right)^{-1} \left(\alpha{\bm X}^*\hspace{-0.2mm}+\hspace{-0.2mm}\beta{\hat {\bm L}}{\hat {\bm A}}{\bm X}^*+\gamma{\hat {\bm L}} {\bm X}^*\right),
\end{aligned}
\label{energy8}
\end{equation} 
which recovers the node global embeddings \( \bm H^{*} \) in Eq.~\eqref{ActiveD8}.
This recovery indicates that the global embeddings of nodes can serve as the optimal solution to the energy function and can be considered high-quality node representations.

Building on the findings in Section~\ref{Active Diffusion on Graphs} regarding the global embeddings of nodes, we can conclude that these embeddings are not only stable and capable of capturing comprehensive global structural information but also provide robust and high-quality node representations. These advantages highlight the superiority of our ADGNN in overcoming the over-smoothing issue while efficiently integrating global graph information.

\subsection{Loss Function}
\label{LF}

To obtain comprehensive node representations, we concatenate the ego, local, and global embeddings of nodes as follows:
\begin{equation}
\overrightarrow{\bm H} = \text{Concat}\left(\bm H^{(0)}, \bm H^{(K)}, \bm H^{*}\right),
\label{LOSS1}
\end{equation} 
where \( \overrightarrow{\bm H} \in \mathbb{R}^{N \times 3d'} \) represents the high-level node embeddings. Here, \( \bm H^{(K)} \) from Eq. \eqref{ActiveD3} denotes the local embeddings, capturing local structural information, and \( K \) is a positive integer determined via grid-search optimization. These three components, derived from different diffusion scales, collectively enhance the model's generalization capabilities.

Next, we generate a learnable weight vector \( \mathbf{w} \in \mathbb{R}^{1 \times 3d'} \) and apply the Hadamard product with \( \overrightarrow{\bm H} \) to emphasize significant dimensions. The resulting weighted embeddings are then passed through a fully connected (FC) layer and a Softmax function to compute the probability distributions of nodes across different classes:
\begin{equation}
\bm Y = \operatorname{Softmax}\left(\text{ReLU}\left(\overrightarrow{\bm H} \odot \mathbf{w} \right){\bm W}^{\prime\prime} + \mathbf{b}^{\prime\prime}\right),
\label{hadamard}
\end{equation}
where \( {\bm W}^{\prime\prime} \in \mathbb{R}^{3d' \times C} \) is the weight matrix, \( \mathbf{b}^{\prime\prime} \in \mathbb{R}^{1 \times C} \) is the bias, and \( C \) is the number of node classes.

For large-scale graphs, to further enhance the model's generalization, we modify the computation of $\bm Y$ as follows:
\begin{equation}
\bm Y = \operatorname{Softmax}\left(\text{ReLU}\left(\overrightarrow{\bm H} {\bm W}^{\prime\prime\prime} + \mathbf{b}^{\prime\prime\prime} \right)\bm W^{\prime\prime} + \mathbf{b}^{\prime\prime}\right),
\label{hadamard}
\end{equation}
where \( \bm W^{\prime\prime\prime} \in \mathbb{R}^{3d' \times 3d'} \) is the weight matrix, and \( \mathbf{b}^{\prime\prime\prime} \in \mathbb{R}^{1 \times 3d'} \) is the bias.
Additionally, we employ the cross-entropy (CE) loss function to compute the classification loss:
\vspace{-1mm}
\begin{equation}
\label{predictt}
\mathcal{L}_{\operatorname{ce}} = -\sum_{u \in \mathcal{V}_{\text{labeled}}} \sum_{c=1}^{C} \hat{Y}_{u,c} \log(Y_{u,c}),
\end{equation} 
where \( \mathcal{V}_{\text{labeled}} \) is the set of labeled nodes. \( \hat{Y}_{u,c} \) is an indicator variable equal to 1 if node \( u \) belongs to class \( c \), and 0 otherwise. \( Y_{u,c} \) is the predicted probability of node \( u \) belonging to class \( c \).

\subsection{Computational Complexity}
\label{Computational Complexity}

Let \( N \) denote the number of nodes and \( d \) the initial feature dimension per node. \( d' \) represents the dimensionality of the ego embeddings, while \( C \) is the number of node classes. Additionally, \( K \) specifies the number of diffusion iterations used to obtain local embeddings. With these definitions, a 2-layer GCN has a computational complexity of \( \mathcal{O}(|\mathcal{E}| \cdot d' \cdot (d + C)) \), where \( |\mathcal{E}| \) is the number of edges.

In our ADGNN model, the matrices \( \hat{\bm A} \), \( \hat{\bm L} \), \( \hat{\bm L} \hat{\bm A} \), and \( (\bm I - \delta \hat{\bm A})^{-1} \) are precomputed before the training stage. Each high-level node representation in ADGNN integrates ego, local, and global embeddings. The computational complexities for each component are as follows:
{Ego Embeddings}: \( \mathcal{O}(N \cdot d \cdot d') \);
{Local Embeddings}: \( \mathcal{O}(3K \cdot |\mathcal{E}| \cdot d') \), accounting for \( K \) diffusion iterations;
{Global Embeddings}: \( \mathcal{O}(3N^{2} \cdot d') \);
{Hadamard Product}: \( \mathcal{O}(3N \cdot d') \);
{Prediction Layer}: \( \mathcal{O}(3N \cdot d' \cdot C) \).
Therefore, the total computational complexity of ADGNN is:
\(\mathcal{O}\left(N \cdot d' \cdot (d + 3 + 3C) + 3d' \cdot (|\mathcal{E}| \cdot K + N^{2})\right)
\)

When applied to large-scale graphs, computing \( (\bm I - \delta \hat{\bm A})^{-1} \) becomes computationally intensive. To mitigate this, we employ a Neumann series approximation strategy \cite{NeumannSeries}, as shown below:
\begin{equation}
(\bm I - \delta \hat{\bm A})^{-1} \approx \sum_{t=0}^{T} (\delta \hat{\bm A})^t.
\label{NeumannSeries}
\end{equation}
where \( T \) is the truncation point of the series, which controls the approximation accuracy. As a result, the computational complexity for computing the global embeddings is reduced to \( \mathcal{O}(T \cdot |\mathcal{E}| \cdot d') \).
Consequently, the total computational complexity of our ADGNN model is:
\(\mathcal{O}\left(N \cdot d' \cdot (d + 9d' + 3c) + T \cdot |\mathcal{E}| \cdot d'\right)\).
This complexity is on par with that of a 2-layered GCN, ensuring that ADGNN remains efficient for processing graphs of large scales.

\section{Experimental}
\label{Experimental}

We apply ADGNN to various tasks for evaluation:
1) Node classification on graphs with varying levels of homophily;
2) Pixel classification on a hyperspectral image (HSI);
3) Node classification on large-scale graphs.
In each case, we compare a different set of competing models closely associated with ADGNN and specifically designed for the particular task. 

\begin{table*}[t]
\footnotesize
    \centering
    \caption{Test accuracy of ADGNN and competitors on homophilic and heterophilic graphs. The best results for each dataset are in bold.}
    \vspace{-0.2cm}
    \tabcolsep=3.1pt
    \label{smalldatasets}
    \begin{tabular}{cccccccccc}
    \toprule
Dataset& Cora & Citeseer &Pubmed&Chameleon&Squirrel &Cornell&Texas& Wisconsin\\ 
\midrule 
Type& \multicolumn{3}{c}{Citation networks}& \multicolumn{2}{c}{Wikipedia networks}&  \multicolumn{3}{c}{WebKB networks}\\ 
Homophily ratio & 0.81 &0.74&0.8&0.23&0.22&0.3&0.11&0.21\\
\# Nodes &2,708&3,327&19,717&2,277&5,201&183&183&251\\
\# Edges &5,429&4,732&44,338&31,421&198,493&295&309&499\\
\# Node features &1,433&3,703&500&2,325&2,089&1,703&1,703&1,703\\
\# Classes &7&6&3&5&5&5&5&5\\
    \midrule 
    MLP & 74.8$\pm$2.2  & 72.4$\pm$2.2 &86.7$\pm$0.4 &49.0$\pm$2.4 & 32.9$\pm$1.8&81.1$\pm$6.4 &81.9$\pm$4.8 &85.3$\pm$3.6\\
\midrule 
GCN & 87.3$\pm$1.3 &76.7$\pm$1.6 &87.4$\pm$0.7&60.5$\pm$3.6 &36.9$\pm$1.4 &58.5$\pm$5.0 &59.5$\pm$5.3 &59.8$\pm$6.9 \\
GAT & 87.7$\pm$1.9 &75.5$\pm$1.7 &86.7$\pm$0.5 &61.1$\pm$2.5 &30.6$\pm$2.1 &60.8$\pm$5.2 &58.4$\pm$4.5 &55.3$\pm$8.7 \\
MixHop&87.6$\pm$1.2 & 76.3$\pm$1.3 & 87.3$\pm$0.6 & 62.1$\pm$3.4 &43.8$\pm$1.5 &73.5$\pm$6.3 &77.8$\pm$5.7 &75.9$\pm$4.9 \\
Geom-GCN &84.9$\pm$1.2 &77.0$\pm$1.4&88.9$\pm$0.9 &60.9$\pm$2.4 &36.1$\pm$1.3&61.5$\pm$6.1 &67.6$\pm$5.8 &64.1$\pm$7.3 \\
IDGL & 88.7$\pm$1.2 & 76.7$\pm$1.2 & 89.4$\pm$0.4 &61.5$\pm$2.9 &42.6$\pm$2.9 &84.5$\pm$5.4 &84.9$\pm$4.1 &87.2$\pm$5.5 \\
H$_{2}$GCN&86.9$\pm$1.4&77.1$\pm$1.6 &89.4$\pm$0.3 &57.1$\pm$1.6 &36.4$\pm$1.9 &82.2$\pm$4.8 &84.9$\pm$3.8 &86.7$\pm$4.7 \\ 
\midrule 
NodeFormer & 87.8$\pm$0.9 & 77.6$\pm$1.1 & 89.0$\pm$1.0& 52.8$\pm$2.5& 37.3$\pm$2.0 & 75.3$\pm$5.3&  78.7$\pm$4.5& 79.0$\pm$5.1\\
SGFormer&  88.0$\pm$0.6  & 77.8$\pm$0.7 &  89.2$\pm$0.2 &  59.5$\pm$4.7  &   40.7$\pm$2.0 & 66.5$\pm$5.2  & 70.8$\pm$4.0& 72.9$\pm$5.0 \\
DIFFormer  & 88.5$\pm$1.7  & 77.0$\pm$1.3  & 89.1$\pm$0.5 &  61.7$\pm$3.8&   42.6$\pm$3.5& 84.4$\pm$7.3&  85.4$\pm$5.0& 86.5$\pm$4.1\\ 
\midrule 
GDC& 88.9$\pm$0.9 & 77.0$\pm$0.6  & 89.1$\pm$0.6 & 62.3$\pm$2.0 &  40.5$\pm$1.9&78.8$\pm$4.5 & 80.6$\pm$4.6& 82.0$\pm$6.2\\
GRAND & 87.4$\pm$1.0 & 76.6$\pm$1.8  & 88.1$\pm$0.5&  56.7$\pm$3.2& 37.1$\pm$1.5& 74.1$\pm$7.0&  73.7$\pm$7.3& 76.2$\pm$3.6\\
GRAND++ & 87.6$\pm$1.1  & 76.8$\pm$0.9  & 88.2$\pm$0.9  & 61.5$\pm$2.5  & 40.1$\pm$1.8 & 82.6$\pm$5.2 & 84.8$\pm$5.5& 84.7$\pm$4.8  \\ 
\midrule
\bf{ADGNN}  & \bf{90.0$\pm$1.0} &\bf{78.8$\pm$0.7} & \bf{89.7$\pm$0.4} & \bf{66.0$\pm$2.8}  & \bf{51.4$\pm$1.2}  &  \bf{88.9$\pm$4.9}  &  \bf{90.7$\pm$3.2}  &\bf{92.2$\pm$3.3}  \\
\bottomrule
\end{tabular} 
\end{table*}

\begin{figure*}  
\centering 
\subfigure{\includegraphics[width=7in,height=0.21in]{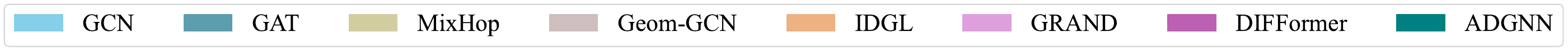}} \\
\vspace{-4mm}
\subfigure{\includegraphics[width=7in,height=1.35in]{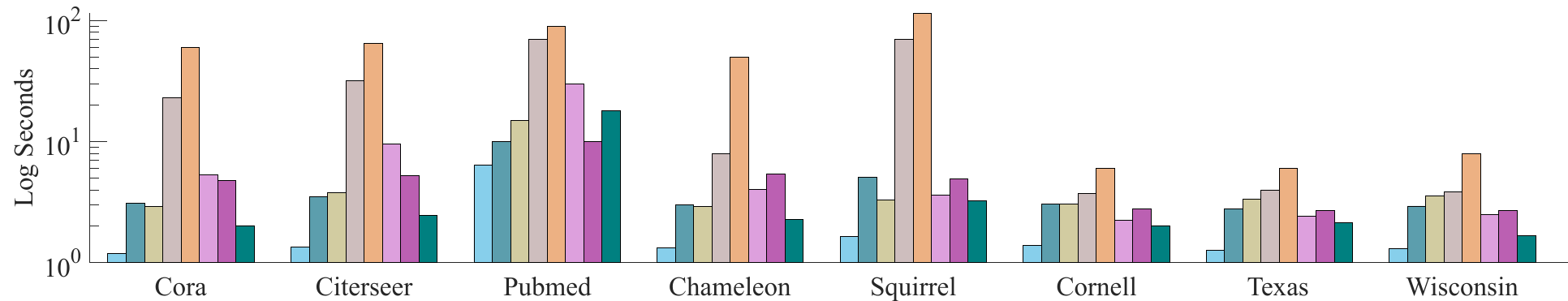}}  
\vspace{-5mm} 
\caption{Running time comparison. Seven competitors and our ADGNN run for 500 epochs during training.}
\label{runtime}
\end{figure*}

\subsection{Node Classification on Graphs with Varying Levels of Homophily}\label{Small} 
To validate the perception and fitting capabilities of ADGNN in complex graph-structured data, we choose eight open graph datasets with varying levels of homophily. 
Table~\ref{smalldatasets} records the statistics information of these graph datasets.
We compare ADGNN with several prominent classification methods:
1) basic model: MLP \cite{MLP};
2) GNN models: GCN \cite{GCN+chapter2017}, GAT \cite{GAT+chapter2018}, 
MixHop \cite{MixHop}, Geom-GCN \cite{Geom+chapter2020}, and H$_{2}$GCN \cite{hhgcn};
3) graph transformers: NodeFormer \cite{NodeFormer}, SGFormer \cite{SGFormer}, and DIFFormer \cite{DIFFORMER};
4) diffusion-based graph models: GDC \cite{GDC}, GRAND \cite{GRAND}, and GRAND++ \cite{GRAND++}.

In this experiment, we adopt the H$_{2}$GCN setup \cite{hhgcn}, randomly splitting the nodes in each class into 48\% for training, 32\% for validation, and 20\% for testing. The results summarized in Table~\ref{smalldatasets} report the mean and standard deviation of five runs with different initializations.
Obviously, all classification methods achieve high accuracy in homophilic graphs. However, in heterophilic graphs, performance varies significantly. For example, in WebKB networks with low homophily, GNNs such as GCN, GAT, MixHop, and Geom-GCN perform unsatisfactorily because of their overreliance on local structural information. In contrast, models such as IDGL (which learns graph structures), H$_{2}$GCN (which separates node and neighbor embeddings), and those that do not depend solely on local graph information—such as NodeFormer, SGFormer, DIFFormer, GDC, and GRAND++—show much better performance. Notably, DIFFormer, which controls the diffusion rate, and GRAND++, which incorporates a source term, exhibit exceptional performance.
Nevertheless, our ADGNN consistently outperforms its competitors across different graphs. This success is attributed to ADGNN's integration of diverse source terms, enabling it to capture complex structural information and combine data from different diffusion scales.

This experiment assesses the computational efficiency of our ADGNN. Due to space constraints, we select several representative baselines and compare the runtime required to train 500 epochs for the baselines and our ADGNN. The results are presented in Fig.~\ref{runtime}.
GCN is the fastest model, with ADGNN ranking second, except on the PubMed dataset. 
This can be attributed to ADGNN’s ability to compute global node embeddings in a single step, bypassing the need for extensive diffusion iterations and saving time. However, ADGNN’s computational complexity is sensitive to the number of nodes.
For large-scale graphs, though, ADGNN’s complexity has been optimized to scale with the number of edges, as discussed in Section~\ref{Computational Complexity}, ensuring that ADGNN remains computationally efficient for both medium-sized and large-scale graphs.
In contrast, IDGL and Geom-GCN are slower due to their iterative graph structure learning modules and geometric aggregation schemes.

\begin{figure}[t]  
\centering 
\subfigure{\includegraphics[width=3.4in,height=4.9in]{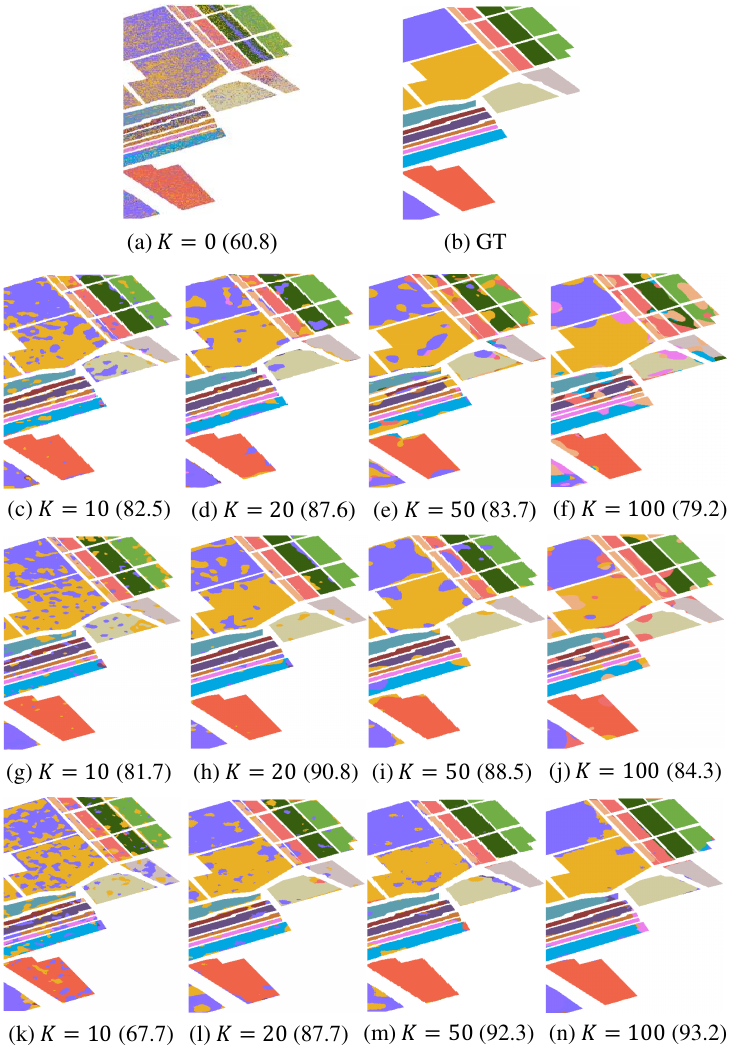}} 
\footnotesize
\vspace{-2mm}
\caption{Pixel classification maps and test accuracies (\%) obtained by GCN, DIFFormer, and our ADGNN across various diffusion iteration numbers $K$ on a real HSI, Salinas. Ground truth (GT) denotes the ideal pixel classification results.
(a) without diffusion, 
(c)-(f) GCN, (g)-(j) DIFFormer, (k)-(n) our ADGNN.}
\label{oversmooth} 
\end{figure}

\subsection{Pixel classification on A Hyperspectral Image}
\label{Long-Range Dependency}
To highlight the advantage of our ADGNN in capturing long-range dependencies and addressing the over-smoothing issue, we apply it to the pixel classification task on a real HSI (Salinas) \cite{salinas}. 
In this context, pixels are treated as nodes, and spatially adjacent pixels are considered neighboring nodes, reflecting the regional continuity typical of natural landscapes and man-made structures.
We randomly choose 3.5\textperthousand~of each class for training, with the remaining data reserved for testing.
We compare ADGNN with two representative competitors: GCN, a classic GNN model that uses current local structural information as node representations and is prone to over-smoothing, and DIFFormer, which mitigates over-smoothing by adjusting the diffusion step size.
To demonstrate the benefits of the active diffusion mechanism, ADGNN uses ego embeddings as the unique source term. Additionally, to examine the effect of the number of diffusion iterations ($K$) on classification performance, ADGNN relies solely on node local embeddings.

Fig.~\ref{oversmooth} presents the experimental results.
Fig.~\ref{oversmooth} (a) shows the classification outcomes for the three methods without any diffusion process. 
Clearly, the diffusion process significantly improves classification accuracy. However, GCN and DIFFormer do not consistently improve with additional diffusion. 
As seen in Fig.~\ref{oversmooth} (f) and (j), a large number of continuous regions are misclassified into the same category, particularly for GCN. 
This indicates that GCN and DIFFormer experience over-smoothing,
especially GCN.
In contrast, our ADGNN demonstrates stable and accurate classification performance.
As shown in Fig.~\ref{oversmooth} (m) and (n), the boundaries between different classes remain clear.
Additionally, test accuracy further confirms the superiority of ADGNN.
These results demonstrate that ADGNN effectively tackles the over-smoothing problem and captures long-range dependencies in hyperspectral image pixel classification.

\begin{table}[t]\footnotesize 
\centering  
\caption{The test accuracy (\%) of ADGNN and competing models, as well as the time consumed (in seconds) for running 100 epochs during the training phase. The best results for each dataset are highlighted in bold. For the ogbn-proteins dataset, the evaluation metric is ROC-AUC.}
\vspace{-2mm}
\tabcolsep=1.2pt
\label{bigbig}
\begin{tabular}{ccccc}
\toprule
  & ogbn-proteins &Amazon2m & Pokec & ogbn-arxiv\\ 
\midrule
\# Nodes&132,534 & 2,449,029 & 1,632,803&169,343\\
\# Edges & 39,561,252&  61,859,140 &30,622,564&1,166,243\\
\# Node features  & 8 & 100 & 65&128\\
\# Classes& 112& 47 & 2&40\\
\midrule 
\multirow{2}{*}{MLP} &70.2$\pm$1.1 &  63.5$\pm$0.1 &  60.1$\pm$0.5 & 55.5$\pm$0.2\\
& 124s    & 426s & 93s & 6s \\
\multirow{2}{*}{GCN} & 72.5$\pm$0.5 &  83.9$\pm$0.1 & 62.3$\pm$1.1  &71.7$\pm$0.3  \\
 &129s   & 485s  & 101s &  8s\\
\multirow{2}{*}{GAT}& 75.2$\pm$1.5 & 85.2$\pm$0.3  &  65.6$\pm$0.3  &  67.6$\pm$0.2\\
 & 183s  & 1578s  & 890s & 180s \\
 \multirow{2}{*}{NodeFormer}  &  77.4$\pm$1.2 &  87.8$\pm$0.2 & 69.3$\pm$0.5
&59.9$\pm$0.4 \\
  & 305s  &   1106s & 290s &  54s \\
\multirow{2}{*}{DIFFormer} &  77.0$\pm$0.6 & 87.2$\pm$0.3 & 69.2$\pm$0.8&72.1$\pm$0.3 \\
 &  146s & 681s & 201s & 40s \\
\multirow{2}{*}{SGFormer} & 77.5$\pm$0.5  &89.1$\pm$0.1&68.5$\pm$0.2 &72.6$\pm$0.1 \\
  & 105s  &  471s &  141s & 10s \\
\multirow{2}{*}{ADGNN}&  {\bf 78.8$\pm$0.5}  & {\bf89.4$\pm$0.1}& {\bf 69.7$\pm$0.3} & {\bf73.3$\pm$0.2}\\
 & 115s & 525s  &  162s & 27s  \\
\bottomrule 
\end{tabular}
\end{table}

\subsection{Node Classification on Large-Scale Graphs}\label{Large}

To assess the generalization ability and computational efficiency of ADGNN, we conduct experiments on four large-scale graph datasets: ogbn-proteins \cite{ogbdataset}, Amazon2m \cite{Amazon2m}, Pokec \cite{DIFFORMER}, and ogbn-arxiv \cite{ogbdataset}. For a fair comparison, we use the train/valid/test splits defined in SGFormer \cite{SGFormer}. Due to the size of these datasets, which poses scalability challenges for many graph models, we compare ADGNN with several computationally efficient baseline models.

To address the memory and computational constraints, we implement mini-batch training with a batch size of $100K$ for both the Amazon2m and Pokec datasets. 
The results recorded in Table~\ref{bigbig} clearly demonstrate that ADGNN consistently outperforms other models, underscoring its strong generalization capability.
Additionally, ADGNN maintains a competitive advantage in computational efficiency, ranking second in training time, behind only MLP, GCN, and SGFormer.
Although ADGNN's computational efficiency is not the highest, it leverages diverse source terms and captures long-range dependencies, which enhances its accuracy, providing it with a notable performance edge.

\vspace{-1mm}
\section{Conclusion}\label{Conclusion}
In this work, we introduce a novel GNN, ADGNN, rooted in active diffusion. Unlike conventional passive-diffusion-based GNNs, ADGNN incorporates three external, informative source terms and extends the diffusion process to infinity. This allows ADGNN to address the over-smoothing problem and effectively capture global structural information.
Moreover, ADGNN completes infinite diffusion iterations in a single step by directly computing the closed-form solution of the diffusion iteration formula. This results in our ADGNN that efficiently captures comprehensive graph information by leveraging diverse source terms and infinite diffusion iterations, leading to more detailed and robust node representations.
Extensive experiments demonstrate that ADGNN outperforms most existing methods in both accuracy and efficiency on node classification tasks across graphs of varying scales and homophily levels.
In future work, we will explore more valuable source terms to enrich the diffusion process, further improving the model's ability to capture complex relationships.

\bibliographystyle{named}
\bibliography{ijcai25}

\begin{thebibliography}{}

\bibitem[\protect\citeauthoryear{Abu{-}El{-}Haija \bgroup \em et al.\egroup }{2019}]{MixHop}
Sami Abu{-}El{-}Haija, Bryan Perozzi, Amol Kapoor, and Nazanin~Alipourfard et~al.
\newblock {MixHop}: Higher-order graph convolutional architectures via sparsified neighborhood mixing.
\newblock In {\em International Conference on Machine Learning}, volume~97, pages 21--29, 2019.

\bibitem[\protect\citeauthoryear{Chamberlain \bgroup \em et al.\egroup }{2021}]{GRAND}
Ben Chamberlain, James Rowbottom, Maria~I. Gorinova, Michael~M. Bronstein, Stefan Webb, and Emanuele Rossi.
\newblock {GRAND:} graph neural diffusion.
\newblock In {\em International Conference on Machine Learning}, volume 139, pages 1407--1418, 2021.

\bibitem[\protect\citeauthoryear{Chen \bgroup \em et al.\egroup }{1987}]{LOG}
Jer{-}Sen Chen, Andres Huertas, and G{\'{e}}rard~G. Medioni.
\newblock Fast convolution with laplacian-of-gaussian masks.
\newblock {\em {IEEE} Transactions on Pattern Analysis and Machine Intelligence}, 9(4):584--590, 1987.

\bibitem[\protect\citeauthoryear{Choi \bgroup \em et al.\egroup }{2023}]{GREAD}
Jeongwhan Choi, Seoyoung Hong, Noseong Park, and Sung{-}Bae Cho.
\newblock {GREAD:} graph neural reaction-diffusion networks.
\newblock In {\em International Conference on Machine Learning}, volume 202, pages 5722--5747, 2023.

\bibitem[\protect\citeauthoryear{Hamilton \bgroup \em et al.\egroup }{2017}]{GraphSage+chapter2017}
William~L. Hamilton, Zhitao Ying, and Jure Leskovec.
\newblock Inductive representation learning on large graphs.
\newblock In {\em Neural Information Processing Systems}, pages 1024--1034, 2017.

\bibitem[\protect\citeauthoryear{Hu \bgroup \em et al.\egroup }{2020}]{ogbdataset}
Weihua Hu, Matthias Fey, Marinka Zitnik, Yuxiao Dong, Hongyu Ren, Bowen Liu, Michele Catasta, and Jure Leskovec.
\newblock Open graph benchmark: Datasets for machine learning on graphs.
\newblock In {\em Neural Information Processing Systems}, 2020.

\bibitem[\protect\citeauthoryear{Jin \bgroup \em et al.\egroup }{2023}]{Amazon2m}
Wei Jin, Haitao Mao, Zheng Li, Haoming Jiang, and Chen~Luo et~al.
\newblock Amazon-m2: {A} multilingual multi-locale shopping session dataset for recommendation and text generation.
\newblock In {\em Neural Information Processing Systems}, 2023.

\bibitem[\protect\citeauthoryear{Kipf and Welling}{2017}]{GCN+chapter2017}
Thomas~N. Kipf and Max Welling.
\newblock Semi-supervised classification with graph convolutional networks.
\newblock In {\em International Conference on Learning Representations}, pages 1--14, 2017.

\bibitem[\protect\citeauthoryear{Klicpera \bgroup \em et al.\egroup }{2019}]{GDC}
Johannes Klicpera, Stefan Wei{\ss}enberger, and Stephan G{\"{u}}nnemann.
\newblock Diffusion improves graph learning.
\newblock In {\em Neural Information Processing Systems}, pages 13333--13345, 2019.

\bibitem[\protect\citeauthoryear{Li \bgroup \em et al.\egroup }{2024}]{HiD-Net}
Yibo Li, Xiao Wang, Hongrui Liu, and Chuan Shi.
\newblock A generalized neural diffusion framework on graphs.
\newblock In {\em {AAAI} Conference on Artificial Intelligence}, pages 8707--8715, 2024.

\bibitem[\protect\citeauthoryear{Liu \bgroup \em et al.\egroup }{2022}]{nlgnn}
Meng Liu, Zhengyang Wang, and Shuiwang Ji.
\newblock Non-local graph neural networks.
\newblock {\em {IEEE} Transactions on Pattern Analysis and Machine Intelligence}, 44(12):10270--10276, 2022.

\bibitem[\protect\citeauthoryear{Miralda-Escud{\'e}}{2003}]{miralda2003dark}
Jordi Miralda-Escud{\'e}.
\newblock The dark age of the universe.
\newblock {\em Science}, 300(5627):1904--1909, 2003.

\bibitem[\protect\citeauthoryear{Pei \bgroup \em et al.\egroup }{2020}]{Geom+chapter2020}
Hongbin Pei, Bingzhe Wei, Kevin~Chen{-}Chuan Chang, Yu~Lei, and Bo~Yang.
\newblock {Geom-GCN}: geometric graph convolutional networks.
\newblock In {\em International Conference on Learning Representations}, 2020.

\bibitem[\protect\citeauthoryear{Su \bgroup \em et al.\egroup }{2023}]{salinas}
Yuanchao Su, Lianru Gao, Mengying Jiang, Antonio Plaza, Xu~Sun, and Bing Zhang.
\newblock {NSCKL:} normalized spectral clustering with kernel-based learning for semisupervised hyperspectral image classification.
\newblock {\em {IEEE} Transactions on Cybernetics}, 53(10):6649--6662, 2023.

\bibitem[\protect\citeauthoryear{Thorpe \bgroup \em et al.\egroup }{2022}]{GRAND++}
Matthew Thorpe, Tan~Minh Nguyen, Hedi Xia, Thomas Strohmer, Andrea~L. Bertozzi, Stanley~J. Osher, and Bao Wang.
\newblock {GRAND++:} graph neural diffusion with {A} source term.
\newblock In {\em International Conference on Learning Representations}, 2022.

\bibitem[\protect\citeauthoryear{Timilsina \bgroup \em et al.\egroup }{2021}]{DIFFUSION0}
Mohan Timilsina, Alejandro Figueroa, Mathieu d'Aquin, and Haixuan Yang.
\newblock Semi-supervised regression using diffusion on graphs.
\newblock {\em Applied Soft Computing}, 104:107188, 2021.

\bibitem[\protect\citeauthoryear{Velickovic \bgroup \em et al.\egroup }{2018}]{GAT+chapter2018}
Petar Velickovic, Guillem Cucurull, Arantxa Casanova, Adriana Romero, Pietro Li{\`{o}}, and Yoshua Bengio.
\newblock Graph attention networks.
\newblock In {\em International Conference on Learning Representations}, 2018.

\bibitem[\protect\citeauthoryear{Wang \bgroup \em et al.\egroup }{2019}]{NeumannSeries}
Fen Wang, Gene Cheung, and Yongchao Wang.
\newblock Low-complexity graph sampling with noise and signal reconstruction via neumann series.
\newblock {\em {IEEE} Trans. Signal Process.}, 67(21):5511--5526, 2019.

\bibitem[\protect\citeauthoryear{Wang \bgroup \em et al.\egroup }{2024}]{PAMI5}
Kun Wang, Yuxuan Liang, Xinglin Li, Guohao Li, and Bernard~Ghanem et~al.
\newblock Brave the wind and the waves: Discovering robust and generalizable graph lottery tickets.
\newblock {\em {IEEE} Transactions on Pattern Analysis and Machine Intelligence}, 46(5):3388--3405, 2024.

\bibitem[\protect\citeauthoryear{Wu \bgroup \em et al.\egroup }{2022a}]{MLP}
Qidie Wu, Jinsheng Kuang, Jiyun Tao, Jienan Chen, and Warren~J. Gross.
\newblock {DsMLP}: a learning-based multi-layer perception for {MIMO} detection implemented by dynamic stochastic computing.
\newblock {\em {IEEE} Transactions on Signal Processing}, 70:6392--6403, 2022.

\bibitem[\protect\citeauthoryear{Wu \bgroup \em et al.\egroup }{2022b}]{NodeFormer}
Qitian Wu, Wentao Zhao, Zenan Li, David~P. Wipf, and Junchi Yan.
\newblock Nodeformer: {A} scalable graph structure learning transformer for node classification.
\newblock In {\em Advances in Neural Information Processing Systems}, 2022.

\bibitem[\protect\citeauthoryear{Wu \bgroup \em et al.\egroup }{2023a}]{DIFFORMER}
Qitian Wu, Chenxiao Yang, Wentao Zhao, Yixuan He, David Wipf, and Junchi Yan.
\newblock {DIFFormer}: Scalable (graph) transformers induced by energy constrained diffusion.
\newblock In {\em International Conference on Learning Representations}, 2023.

\bibitem[\protect\citeauthoryear{Wu \bgroup \em et al.\egroup }{2023b}]{SGFormer}
Qitian Wu, Wentao Zhao, Chenxiao Yang, Hengrui Zhang, Fan Nie, Haitian Jiang, Yatao Bian, and Junchi Yan.
\newblock Simplifying and empowering transformers for large-graph representations.
\newblock In {\em Advances in Neural Information Processing Systems}, 2023.

\bibitem[\protect\citeauthoryear{Xu \bgroup \em et al.\egroup }{2019}]{GIN+chapter2019}
Keyulu Xu, Weihua Hu, Jure Leskovec, and Stefanie Jegelka.
\newblock How powerful are graph neural networks.
\newblock In {\em International Conference on Learning Representations}, pages 1--17, 2019.

\bibitem[\protect\citeauthoryear{Zhang \bgroup \em et al.\egroup }{2023}]{PAMI3}
Rui Zhang, Yunxing Zhang, Chengjun Lu, and Xuelong Li.
\newblock Unsupervised graph embedding via adaptive graph learning.
\newblock {\em {IEEE} Transactions on Pattern Analysis and Machine Intelligence}, 45(4):5329--5336, 2023.

\bibitem[\protect\citeauthoryear{Zhao \bgroup \em et al.\egroup }{2021}]{nips2}
Jialin Zhao, Yuxiao Dong, Ming Ding, Evgeny Kharlamov, and Jie Tang.
\newblock Adaptive diffusion in graph neural networks.
\newblock In {\em Neural Information Processing Systems}, pages 23321--23333, 2021.

\bibitem[\protect\citeauthoryear{Zhu \bgroup \em et al.\egroup }{2020}]{hhgcn}
Jiong Zhu, Yujun Yan, Lingxiao Zhao, Mark Heimann, Leman Akoglu, and Danai Koutra.
\newblock Beyond homophily in graph neural networks: current limitations and effective designs.
\newblock In {\em Neural Information Processing Systems}, 2020.

\end{thebibliography}

\end{document}